\documentclass[runningheads]{llncs}

\usepackage{hyperref}

\usepackage{float} 
\usepackage{multirow} 
\usepackage{booktabs} 
\usepackage{amsmath} 
\usepackage{slashbox} 
\usepackage[T1]{fontenc}  
\usepackage{textcomp}     

\usepackage{subcaption}
\usepackage{fancyvrb} 
\usepackage{adjustbox} 

\usepackage[color]{changebar} 
\usepackage{wrapfig}

\begin{document}
\title{Identifying the Units of Measurement in Tabular Data}

\author{Taha Ceritli\inst{1,2}\and\\
Christopher K.~I.~Williams\inst{1,2}}
\institute{
School of Informatics, University of Edinburgh, UK 
\and 
Alan Turing Institute, London, UK
}
\maketitle             
\begin{abstract}
We consider the problem of identifying the units of measurement
in a data column that contains both numeric values and unit
symbols in each row, e.g., ``5.2 l'', ``7 pints''. In this case
we seek to identify the dimension of the column (e.g.\ volume) and
relate the unit symbols to valid units (e.g.\ litre, pint)
obtained from a knowledge graph. Below
we present \emph{PUC}, a Probabilistic Unit Canonicalizer that
can accurately identify the units of measurement, extract semantic
descriptions of quantitative data columns and canonicalize their
entries. We present the first messy real-world tabular datasets
annotated for units of measurement, which can enable and accelerate
the research in this area. Our experiments on these datasets show that
PUC achieves better results than existing solutions.

\end{abstract}

\section{Introduction}
\label{sec:introduction}
\makeatletter{\renewcommand*{\@makefnmark}{}
\footnotetext{Presented at the ECML-PKDD Workshop on Automating Data Science, 17 Sept 2021.}\makeatother}
In a typical data analytics project, a large amount of time and effort is spent on understanding and cleaning the data. A particular task, which is often performed manually, is to understand how measurements are encoded in quantitative data columns. A \emph{measurement} is a combination of a numeric value and a unit (e.g., litre), which can be encoded by several possible unit symbols for that unit (e.g., \texttt{l} and \texttt{L}). 

\begin{figure}[H]
  \centering
    \includegraphics[width=.7\textwidth]{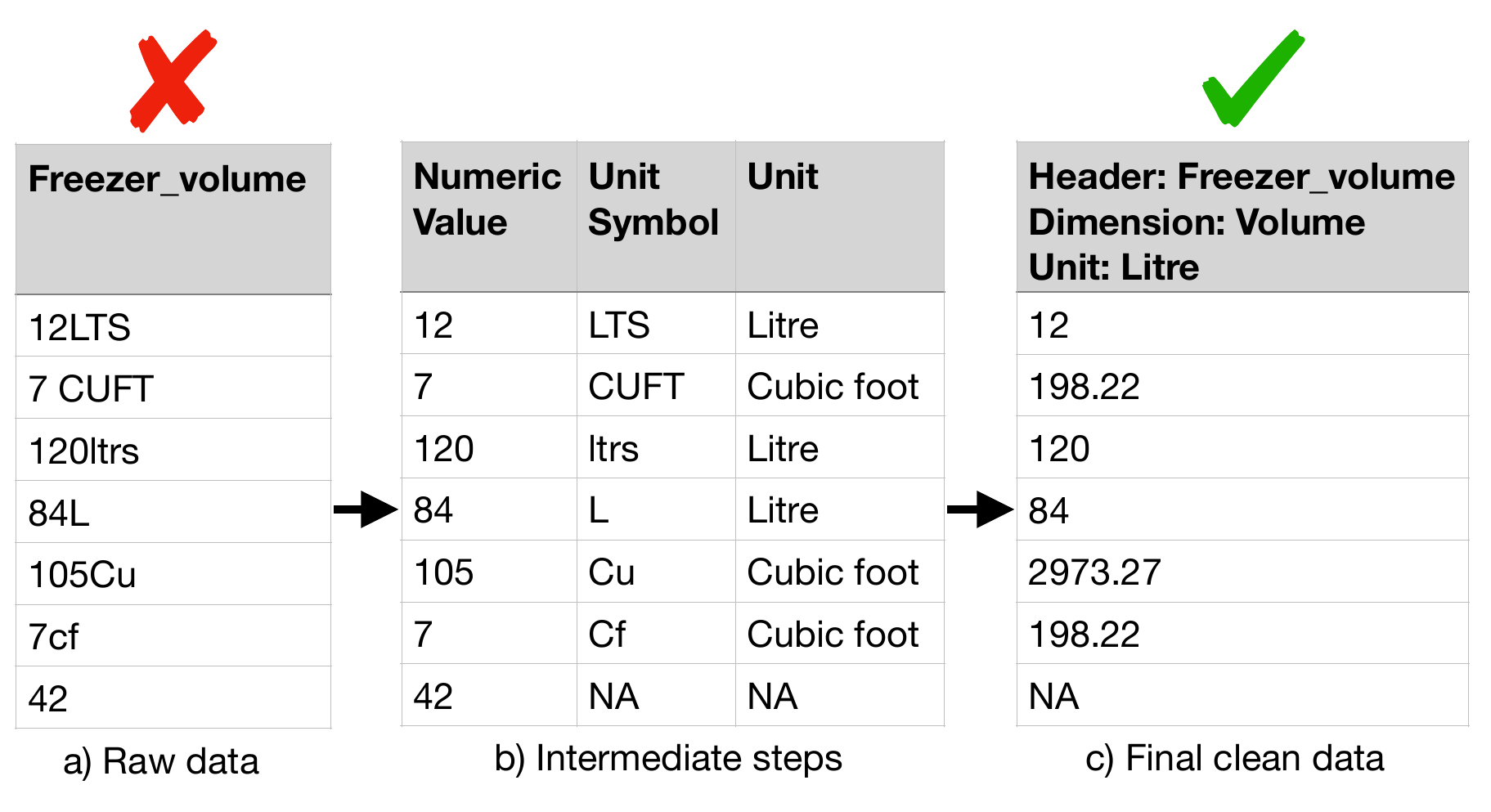}
  \caption{A motivating real-world example that represents our pipeline. a) shows the samples of a raw dataset. b) indicates the intermediate steps required to transform the column. c) denotes the final data column obtained by applying the transformations.}
\label{fig:motivation}
\end{figure}

Consider the following example: Suppose we are given the quantitative data column\footnote{The data is sampled from the ``Freezer\_volume'' column of the Household Electricity Survey (HES) dataset which can be accessed by registering at \url{https://tinyurl.com/ybbqu3n3}.
} in Figure \ref{fig:motivation}(a), which denotes the volume of freezers in various households. The value of each data entry is a measurement encoded by a unit symbol, except the last entry which does not have any unit symbol (we refer to the absence of a unit symbol in an entry as the \emph{missing unit}). As shown in Figure \ref{fig:motivation}(b), the measurements are encoded with two distinct units (litres and cubic feet) and six different unit symbols, some of which including \texttt{ltrs}, and \texttt{Cu} can be considered as \emph{anomalous unit symbols} as they do not follow the standard encodings of units. Moreover, a unit (e.g., litre) can be informative about the \emph{dimension} of a measurement (in this case volume), which is defined as an expression of the dependence of a physical quantity on mutually independent components called \emph{base quantities} in the International Vocabulary of Metrology\footnote{The document is accessible at \url{https://www.bipm.org/documents/20126/2071204/JCGM_200_2012.pdf/f0e1ad45-d337-bbeb-53a6-15fe649d0ff1}}.

We define \emph{Unit Canonicalization (UC)} as the task of identifying the dimension of a data column and the units of all its entries so that the data entries can be canonicalized (i.e., so they all are expressed in the same units). See Figure \ref{fig:motivation}(c) for an example. To address the UC problem, one needs to apply transformations such as parsing and identifying the units in the entries, inferring the common unit for the column which is placed in the metadata (e.g., the header), making the entries numeric and scaling the entries where needed. To the best of our knowledge, the UC problem is not addressed by any existing work in the literature (see Sec.\ \ref{sec:related_work} for a detailed discussion). In this work, we propose a probabilistic approach to the UC problem. Our contributions are as follows:
\begin{itemize}
\item We propose a probabilistic model which allows us to extract semantic information about a given data column containing unit symbols (such as its dimension and unit) and canonicalize its entries (Section \ref{sec:methodology}).
\item We make the first quantitative comparison of the existing methods on the unit identification task in real-world tabular data (Section \ref{sec:experiments}).
\item We present the first set of real-world datasets annotated for the units of measurement, to accelerate research in this area.
\end{itemize}

\section{Methodology}
\label{sec:methodology}

\paragraph{\bf{Representing Units:}} We represent units by extending a dictionary of units curated from Wikipedia\footnote{Accessible at \url{https://github.com/marcolagi/quantulum/blob/master/quantulum/units.json}} with information from WikiData \cite{vrandecic2014} and QUDT (Quantities, Units, Dimensions and Data Types Ontology)\footnote{https://www.qudt.org/}. We associate each unit with a dimension and a list of unit symbols. See Appendix \ref{sec:appendix-ontology} for a detailed description.

\paragraph{\bf{The Proposed Model:}} We assume that a column of data $\textbf{y} = \{y_i\}_{i=1}^N$ consisting of $N$ rows has been read in, where each $y_i$ denotes the characters in the $i^{th}$ row. Each $y_i$ is parsed using regular expressions (see Appendix \ref{sec:appendix-parsing}) to a numeric value $v_i$ and a unit symbol $x_i$, which may be missing for some entries, i.e., $x_i$ may be null. Our focus here is on the unit symbol
$x_i$ obtained. Denoting the number of possible dimensions for a column by $K$, our model has the following generative process:
\begin{eqnarray*}
\text{column dimension $t$} &\sim& \mathcal{U}(1,K), \nonumber \\
\text{row unit $u_i$} &\sim& p(u_i|t), \nonumber \\
\text{row label $z_i$}  &=&
\begin{cases} 
\text{u}_{\text{i}} & \text{with probability $w_{u_i}^{u_i}$}\ ,\\
\text{m} & \text{with probability $w_{u_i}^m$},\\
\text{a} & \text{with probability $w_{u_i}^a$}\ ,\\
\end{cases} \nonumber\\
\text{row symbol $x_i$} &\sim& p(x_i | z_i),
\end{eqnarray*}
where $\mathcal{U}$ denotes a discrete Uniform
distribution, and $p(u_i|t)$ denotes how likely it is that a row unit $u_i$ represents the dimension $t$. We model $p(u_i|t)$ with an indicator function that assigns a non-zero score only when a unit is a known unit of a dimension. Each row label $z_i$ can be either the same as $u_i$, $m$ or $a$ according to the mixing proportions denoted by $\boldsymbol{W}$, where $m$ and $a$ respectively denote a missing or anomalous unit. Here the mixing proportions $w_{u_i}^{u_i}+ w_{u_i}^m + w_{u_i}^a = 1$ for each row unit $u_i$. Since entries are often expected to be of a regular row label rather than the missing or anomalous labels, we favour regular labels during inference by using lower coefficients for the missing and anomalous labels, i.e. $w_{u_i}^m < w_{u_i}^{u_i}$ and $w_{u_i}^a < w_{u_i}^{u_i}$. These mixing proportions $\boldsymbol{W}$ are assumed to be fixed and known. Even though one could also learn the mixing proportions, this may not be vital as long as the coefficients of the regular labels are larger than the others. Finally, $p(x_i|z_i)$ denotes our observation model which is explained below. Note that the above model is based on ptype \cite{ceritli2020ptype}, but extended to address unit canonicalization by incorporating semantic information 
about the units of measurements.

We build the observation model $p(x_i | z_i)$ upon three functions. First, we develop a Categorical distribution for row unit $u_i$ where the categories correspond to the possible unit symbols for that unit, i.e., $p(x_i | z_i = u_i) = \mathcal{C}at(x_i, \pi_{u_i})$ where $\sum_{s=1}^{S_{u_i}} \pi_{u_i}^s = 1$. Here, $\pi_{u_i}^s$ denotes the probability of observing the $s^{th}$ unit symbol for a unit $u_i$, while $S_{u_i}$ denotes the number of known unit symbols for that unit. Secondly, we model missing units with an indicator function, which assigns a non-zero probability only when a unit symbol is missing. Lastly, we adapt the anomaly type in ptype, which is built based on the idea of an \emph{X-factor} proposed by \cite{quinn2009factorial}, to model anomalous unit symbols. Here, we introduce a likelihood function that assigns low but non-zero probabilities to any data value, which in turn allows the model to detect anomalous unit symbols which do not fit any known unit. Note that our model is fed with unit symbols, which are parsed using regular expressions. See Appendix \ref{sec:appendix-parsing} for the details about our parsing step.

\paragraph{\bf{Inference:}} Given the row symbols $\textbf{x}$ in a data column, the initial task is to infer the column dimension $t$, which is cast to as the problem of calculating the posterior distribution of $t$ given $\textbf{x}$, namely $p(t|\textbf{x})$. We then compute a posterior distribution over each row label conditioned on the dimension and the observed value, i.e., $p(z_i|t,x_i)$. Next, we determine the row units by calculating the posterior distribution of each row unit $u_i$ given $t$, $z_i$ and $x_i$, which is also used to predict the column unit. Lastly, we map a unit symbol labelled as anomalous to a known unit symbol and perform unit canonicalization. The derivations are presented for reproducibility in Appendix \ref{sec:appendix-derivations}.

\section{Related Work}
\label{sec:related_work}
We are not aware of existing work specifically on the unit canonicalization problem. The closest related works are described below:

\cite{van2010converting,hignette2009fuzzy,samadian2014automatic} use semantic web technologies to annotate quantitative data columns in terms of units. However, their annotations are based on ontological classes in domain-specific ontologies rather than the dimensions considered in this work. For example, \cite{samadian2014automatic} define the ``High-Systolic-Blood-Pressure-Measurement'' class with \texttt{kilopascal} as its unit. Given a data column and its ontological class, their goal is to represent all entries with the same predefined unit of the corresponding ontological class. \cite{chambers2010reasoning} consider the task of annotating data entries in spreadsheets rather than tabular data. The authors split the label in a header (e.g., ``Total Gallons'') into separate words (e.g., ``Total'' and ``Gallons''), remove word inflections (e.g., obtaining ``Gallon'' from ``Gallons'') and map word stems into known units and dimensions (e.g., ``Gallon'' and ``volume''). Although their approach can be useful when information is explicitly given in a label, it would not work when the information is given implicitly, as in the label of ``Credit Card Charges'' which implicitly implies that the dimension is currency.

Regular expressions have been used for parsing and identifying units in text and unit conversion \cite{pint2019,wolfram2019alpha}. However, they are designed neither to annotate quantitative data columns with dimensions nor to canonicalize the units of their entries. They do not even take the input as a data column, except for the professional version of Wolfram$|$Alpha which is not freely available. \cite{shbita2019parsing} develop a rule-based system named CCUT which maps unit symbols in tabular data to an ontology so that the data entries can be annotated with semantic information, which can be useful for table understanding. Unlike us, the authors do not use the contextual information in the entries of a data column, i.e., the unit symbols in the entries of a data column may be related to each other through the column dimension. Moreover, they do not consider the task of annotating quantitative data columns with dimensions.

Quantulum \cite{quantulum2016} use an ML model to disambiguate unit symbols in unstructured text (e.g., whether ``pound'' in a sentence refers to currency or mass). However, the entries of a data column consist of only numeric values and unit symbols, unlike long sentences where additional information is available through the other words. Similarly, \cite{foppiano2019automatic} propose an ML framework named Grobid-Quantities (GQ) for processing text documents such as PDF files. However, identifying units in general tabular data can be more challenging since units in the scientific papers are more likely to follow standard notations. \cite{williams2020units} apply simple ML techniques for dimension inference (e.g., the cosine similarities between word-embeddings representations of the header and the units), and then assign the most frequent unit for the inferred dimension as the column unit. Although this approach can be useful for dimension inference, it may be misleading for column unit inference as less frequent units can also be used to encode measurements in data columns. Lastly, dimension inference could benefit from Named Entity Recognition (NER) models. However, the set of tags supported by existing NER models are typically limited, e.g., Stanford NER includes only currency and time\footnote{See \url{https://nlp.stanford.edu/software/CRF-NER.shtml} for a complete list of the \\supported tags.}.

\section{Experiments}
\label{sec:experiments}

\paragraph{\bf{Experimental Setup:}} Here, we briefly describe the datasets, baseline methods and evaluation metrics used in our experiments. See Appendix \ref{sec:appendix-exp-setup} for a detailed description. The datasets, their sources and our annotations for dimensions and units can be accessed via \url{https://tinyurl.com/ay44c62r}. The code to reproduce the experiments are publicly available at \url{https://github.com/tahaceritli/puc}.

We conduct experiments on 24 data columns obtained from 16 CSV data files, each of which contains at least one column where the measurements are encoded by units of measurement. Their dimensions were annotated by hand for these sets, resulting in 2 currency, 2 data storage, 6 mass, 3 volume and 11 length columns. We also annotated each data entry in terms of its numeric value and unit symbol. We construct baselines for dimension inference by adapting CCUT, Grobid-Quantities (GQ), Pint, Stanford NER (S-NER) and Quantulum. On dimension inference, we compare our method with CCUT, GQ, Pint and Quantulum. To assess dimension inference, we use the overall accuracy and the Jaccard index. See \cite{ceritli2020ptype} for a detailed description of these metrics and how they are used for type inference. Additionally, we evaluate the runtime of each method per data column (see Figure \ref{fig:units-runtimes} in Supp.\ Mat.). To measure the performance on the unit identification task, we report the accuracies of the methods per dataset and apply paired t-tests to determine whether the predictions of the competitor methods are significantly different from the predictions of our method. 

\paragraph{\bf{Results:}} Table \ref{table:units-dimensions-explicit-metrics} presents the performance of the methods on the column dimension inference task. The overall accuracies show that PUC performs better than the competitor methods. We observe a similar trend with the performance per dimension, quantified through the Jaccard index. These improvements are due to our model's extensive knowledge about units, and its structure that takes into account the context shared among data rows. Note that the Jaccard index becomes zero when a method incorrectly labels all the data columns of a particular dimension (e.g., CCUT, GQ and Pint lead to zero Jaccard index for the currency dimension because they do not support currency symbols such as \texttt{\$} and \texttt{£} which occur in the entries of the ``Currency" column of the Zomato dataset).

\vspace*{-5mm}
\begin{table}[ht!]
\centering
\scalebox{0.65}{
\large
\begin{tabular}{l|cccccc}
\toprule
\backslashbox{\textbf{Dimension}}{\textbf{Method}} &  \textbf{CCUT} &  \textbf{GQ} & \textbf{Pint} &   \textbf{S-NER} &  \textbf{Quantulum} & \textbf{PUC} \\
\midrule
Currency&  0.00 &    0.00 &  0.00 & 0.67 &      0.50 &        \textbf{1.00} \\
Data storage&  0.00 &    0.00 &  0.00 &  0.00 &      0.50 &        \textbf{1.00} \\
Length&  0.27 &    0.45 &  0.67 &    0.62 & 0.58 &        \textbf{0.91} \\
Mass&  0.00 &    0.67 &  0.57 &   0.67    & \textbf{1.00} &        \textbf{1.00} \\
Volume&  0.00 &    0.00 &  0.00 &   0.33 & 0.67 &   \textbf{1.00} \\
\bottomrule
\addlinespace[1mm]
Overall & \multirow{2}{*}{0.12} &  \multirow{2}{*}{0.38} &  \multirow{2}{*}{0.50} &    \multirow{2}{*}{0.71}  & \multirow{2}{*}{0.71} &   \multirow{2}{*}{\bfseries  0.96} \\
Accuracy &&&&& \\
\hline 
\end{tabular}
}
\vspace*{3mm}
\caption{Performance of the methods for dimension inference using the Jaccard index
and overall accuracy. We highlight the best score in each row by making the highest score bold.}
\label{table:units-dimensions-explicit-metrics}
\end{table}
\vspace*{-10mm}

\begin{wraptable}{r}{5.5cm}
\vspace*{-8mm}
\centering
\scalebox{0.77}{
\begin{tabular}{l|ccccc}
\toprule
\textbf{Dataset} & \textbf{CCUT}&  \textbf{GQ} & \textbf{Pint}&  \textbf{Quant.} &  \textbf{PUC} \\
\midrule
Arabica &  0.77 &               0.17 &  0.66 &      \textbf{1.00} &                0.70 \\
HES &  0.18 &               0.00 &  0.27 &       0.64 &                \textbf{0.98} \\
Huffman                          &  0.06 &               0.53 &  \textbf{1.00} &       \textbf{1.00} &               \textbf{1.00} \\
Maize &  1.00 &               0.30 &  \textbf{1.00} &       \textbf{1.00} &                \textbf{1.00} \\
MBA &  0.00 &               0.00 &  0.00 &       0.84 &                \textbf{0.95} \\
Open Units                       &  0.00 &               0.00 &  \textbf{1.00} &       0.00 &                \textbf{1.00} \\
PHM      &  0.99 &               0.95 &  0.99 &       \textbf{1.00} &                \textbf{1.00} \\
query\_2                          &  0.00 &               0.00 & \textbf{1.00} &       0.00 &                \textbf{1.00} \\
query\_4                          &  0.00 &               0.00 &  \textbf{1.00} &       0.00 &                \textbf{1.00} \\
Robusta &  0.71 &               0.43 &  \textbf{1.00} &       \textbf{1.00} &                \textbf{1.00} \\
Zomato &  0.00 &               0.00 &  0.00 &       0.00 &                \textbf{0.60} \\
143\dots23                     &  0.00 &               0.69 &  0.00 &       0.00 &                \textbf{0.97} \\
143\dots62                    &  0.58 &               0.00 &  0.00 &      \textbf{0.95} &                \textbf{0.95} \\
228\dots96   & \textbf{1.00} &               0.00 &  \textbf{1.00} &      \textbf{1.00} &                \textbf{1.00} \\
3b5\dots ff &  0.60 &               0.00 &  0.00 &      \textbf{1.00} &                \textbf{1.00} \\
\bottomrule
\addlinespace[1mm] 
Overall & \multirow{2}{*}{0.39} &  \multirow{2}{*}{0.20} &  \multirow{2}{*}{0.59} &       \multirow{2}{*}{0.62} &   \multirow{2}{*}{\bfseries  0.94} \\
Accuracy &&&&& \\
\hline 
\end{tabular}
}
\vspace*{3mm}
\caption{Accuracy of the methods on unit identification. We highlight the best score in each row by making the highest score bold. Quantulum is abbreviated as Quant.}
\label{table:accuracy-unit-identification}

\end{wraptable} 
{Table \ref{table:accuracy-unit-identification} presents the accuracy of each method on each dataset (aggregated over columns) and the overall accuracy (averaging over all datasets). PUC performs consistently better than the competitor methods, although they are competitive with our method on some of the datasets. The performance of PUC on the Arabica and Zomato datasets are discussed in Appendix \ref{sec:app-exp-res}. Note that accuracy becomes zero when a method fails to identify the unit of any data entry of a dataset (e.g., Pint cannot recognize the unit symbols in the MBA dataset such as \texttt{LB} and \texttt{OZ} which respectively denote the pound and ounce units). Finally, paired t-tests confirm that PUC is significantly different than each competitor method at the 0.05 level.

\clearpage
\bibliographystyle{splncs04}
\bibliography{references}

\clearpage
\appendix
\begin{center}
\textbf{\huge Supplemental Materials}
\end{center}
Below, we give additional information about our knowledge graph of units (Appendix \ref{sec:appendix-ontology}), describe our experimental setup (Appendix \ref{sec:appendix-exp-setup}), provide additional information about our method (Appendix \ref{sec:app-method}) and discuss our experimental results further (Appendix \ref{sec:app-exp-res}). 

\section{Additional Information about Our Knowledge Graph }
\label{sec:appendix-ontology}

We represent units by extending a dictionary of units curated from Wikipedia\footnote{Accessible at \url{https://github.com/marcolagi/quantulum/blob/master/quantulum/units.json}} with information from WikiData \cite{vrandecic2014} and QUDT (Quantities, Units, Dimensions and Data Types Ontology)\footnote{https://www.qudt.org/}. \cite{keil2019comparison} show that WikiData is the most comprehensive knowledge graph for units and that QUDT contains additional information to WikiData. By extending the existing dictionary, we increase the number of units from 284 to 1080 (we only consider the units in English, although extensions to other languages are straightforward). 

Table \ref{table:unit-ontology} presents two instances that respectively represent the units of litre and gram. The instances in the original dictionary can have six attributes: name, surfaces, entity, URI, dimensions, and symbols. Note that the ``surfaces'' attribute denotes a list of \emph{strings} that refer to a unit, whereas the ``symbols'' attribute is a list of possible symbols and abbreviations for that unit. 

\begin{table}[H]
\centering
\large
\scalebox{0.65}{
\begin{tabular}{llllll}
\toprule
  \textbf{name} & \textbf{surfaces} & \textbf{entity} & \textbf{URI} & \textbf{dimensions} & \textbf{symbols} \\
\midrule
\multirow{2}{*}{litre} &  cubic decimetre, litre, &  volume &  .../wiki/Litre &  \{'base': 'decimetre',  &  l, L, ltr \\
 & cubic decimeter, liter & & &'power': 3\} &\\
  gram & gram, gramme &    mass &  .../wiki/Gram & \textemdash & g, gm \\
\bottomrule
\end{tabular}
}
\vspace*{3mm}
\caption{Two elements of the unit dictionary. Note that URIs begin with \url{https://en.wikipedia.org/}.}
\label{table:unit-ontology}
\end{table}

For each instance we extract a dimension, a unit and a list of unit symbols. The name and entity attributes of the instances in Table \ref{table:unit-ontology} are respectively used as units and their dimensions. Note that the naming convention used in the dictionary differs from our terminology in that it refers to dimension as entity and uses the ``dimensions" attribute to encode the relationship between units. The ``surfaces'' and ``symbols'' attributes are combined to build a set of unit symbols. To search for additional symbols of a unit, we query WikiData using the URI attribute. Additionally, the QUDT reference ID, when available in the response obtained from WikiData, is used to query QUDT. 

We support the following dimensions: acceleration, amount of substance, angle, area, capacitance, catalytic activity, charge, currency, current, data storage, data transfer rate, dimensionless, dynamic viscosity, electric potential, electrical conductance, electrical resistance, energy, flux density, force, frequency, illuminance, inductance, instance frequency, irradiance, kinematic viscosity, length, linear mass density, luminance, luminous flux, luminous intensity, magnetic field, magnetic flux, magnetomotive force, mass, mass flow, power, pressure, radiation absorbed dose, radiation exposure, radioactivity, sound level, speed, temperature, time, torque, typographical element, volume, volume (lumber), volumetric flow.

To query WikiData, Wikipedia and QUDT, we respectively use wikidata \cite{wikidata2017}, wikipedia \cite{wikipedia2016} and pyqudt \cite{pyqudt2019}.

\section{Additional Information about Our Experimental Setup}
\label{sec:appendix-exp-setup}

\subsubsection*{Datasets}
We conduct experiments on 24 data columns obtained from 16 CSV data files, each of which contains at least one column where the measurements are encoded by units of measurement. The datasets used can be briefly described as follows:
\begin{itemize}
\item Arabica, Robusta: a collection of reviews about coffee beans.
\item HES (Household Electricity Survey): time series measurements of the electricity use of domestic appliances (to gain access to the data, please register at \url{https://tinyurl.com/ybbqu3n3}).
\item Huffman: the Huffman Prairie flight trials in 1904, which is available through the U.S.\ Centennial of Flight Commission. 
\item Maize Meal: a list of maize meal products.
\item MBA: a list of products in a grocery shop.
\item Open Units: a list of 1,000 standard servings of branded drinks and their alcohol content.
\item PHM Collection:  information on the collection of Powerhouse Museum Sydney, including textual descriptions, physical, temporal, and spatial data as well as, where possible, thumbnail images.
\item 143\dots6, 143\dots23 and 228\dots96: a set of data Web tables from T2Dv2 Gold Standard to evaluate matching systems on the task of matching Web tables to the DBpedia knowledge base.
\item query\_2, query\_4: a set of tables extracted from WikiData using the properties of height or weight.
\item Zomato: information about restaurants extracted from Zomato.
\end{itemize}

The dimensions of the data columns were annotated by hand for these sets, resulting in 2 currency, 2 data storage, 6 mass, 3 volume and 11 length columns. We also annotated each data entry in terms of its numeric value and unit symbol. Note that numeric values are missing in 5 data columns, i.e., only unit symbols are observed in the entries. For example, the ``Quantity Units" column of the Open Units dataset consists of three unique values. These are \texttt{ml}, \texttt{pint} and \texttt{cl}, which respectively denote the units of millilitres, pint and centiltres. Table \ref{table:units-dataset-scale} presents the number of entries, unique entries, units and unit symbols per data column. The number of unit symbols per data column varies between 2 and 11. 

\begin{table}[ht!]
\centering
\scalebox{0.7}{
\large
\begin{tabular}{llllll}
\toprule
\textbf{dataset} & \textbf{column}&   \textbf{\# non-missing entries} &  \textbf{\# unique entries} &  \textbf{\# units} &  \textbf{\# unit symbols} \\
\midrule
Arabica &            Bag W\dots&       1,283 &                51 &        2 &               2 \\
Arabica &              Altitude &        132 &                81 &        2 &              11 \\
HES &        Freezer\dots&         50 &                47 &        2 &              11 \\
HES&   Refrig\dots&         38 &                37 &        2 &               9 \\
Huffman &              DIST\dots&         71 &                68 &        2 &               2 \\
Maize &            PACK \dots&         30 &                10 &        2 &               2 \\
MBA &  CURR\dots &         43 &                19 &        3 &               6 \\
Open U\dots&        Quanti\dots&       1,082 &                 3 &        3 &               3 \\
PHM &                Height &      30,312 &              1,313 &        3 &               5 \\
PHM &                Weight &        179 &               135 &        2 &               3 \\
PHM &                 Width &      36,330 &              1,282 &        3 &               5 \\
PHM &                 Depth &      19,400 &              1,073 &        3 &               4 \\
PHM &              Diameter &       2,106 &               377 &        2 &               2 \\
Robusta &            Bag W\dots&         28 &                 4 &        2 &               2 \\
Robusta &              Altitude &          8 &                 3 &        1 &               2 \\
query\_2 &       unitH\dots&         22 &                 3 &        3 &               3 \\
query\_2 &        unitW\dots&         22 &                 3 &        3 &               3 \\
query\_4 &       unitH\dots&         22 &                 3 &        3 &               3 \\
Zomato &              currency &       6,386 &                 9 &        9 &               9 \\
143\dots62 &                FORMAT &        123 &                19 &        2 &               3 \\
143\dots23 &                  Size &       3,855 &              1,667 &        2 &               2 \\
228\dots96 &                  Size &        377 &               256 &        2 &               2 \\
3b5\dots ff &                amount &          5 &                 5 &        2 &               2 \\
\bottomrule
\end{tabular}

}
\vspace*{3mm}
\caption{Size of the datasets used and the number of units and unit symbols in each data column.}

\label{table:units-dataset-scale}
\end{table}

\subsubsection*{Baselines}
\label{sec:units-baselines}
As we describe in Sec.\ \ref{sec:related_work}, the methods proposed in \cite{van2010converting,hignette2009fuzzy,samadian2014automatic,shbita2019parsing,foppiano2019automatic,pint2019,quantulum2016,finkel2005incorporating} do not address the unit canonicalization task. However, we can construct baselines to be used in our experiments by adapting some of these methods, namely CCUT \cite{shbita2019parsing}, Grobid-Quantities (GQ) \cite{foppiano2019automatic}, Pint \cite{pint2019}, Stanford NER (S-NER, \cite{finkel2005incorporating}) and Quantulum \cite{quantulum2016}. All these methods take as input a sentence (e.g., ``\dots 2 litres of water.'') and annotate the words that refer to quantities (e.g., 2 litres) with their dimensions (e.g., volume), except S-NER which needs to be trained again for the dimension prediction. Here, we use the data values of the entries as inputs to these methods. 

We construct baselines for dimension inference as follows. We first identify the dimensions of the entries of a data column using a competitor method and then predict the column dimension through majority voting, i.e., assigning the most common dimension as the column dimension. The baseline method based on the S-NER is obtained by training the pre-trained model with pairs of unit symbols and their dimensions (e.g., \texttt{metres} - length). The resulting model then generates a tag for each data entry of a  data column and can be used to assign a dimension to that column through majority voting.

On the unit identification task, we evaluate whether the unit of a data entry can be correctly identified, e.g., \texttt{1 cm} as \texttt{1} and \texttt{centimetre}. For this, we compare our method with CCUT, GQ, Pint and Quantulum, which are developed for identifying units in textual documents. 

\subsubsection*{Evaluation Metrics}
We use different sets of metrics for dimension inference and unit identification. For dimension inference, we use the overall accuracy and the Jaccard index. See \cite{ceritli2020ptype} for a detailed description of these metrics and how they are used for type inference. To measure the performance on the unit identification task, we report the accuracies of the methods per dataset and apply paired t-tests to determine whether the predictions of the competitor methods are significantly different from the predictions of our method. Note that the accuracy on the unit identification task measures the ratio of number of entries in a dataset for which the unit is correctly predicted over the number of entries in a dataset. 

\section{Additional Information about Our Method}
\label{sec:app-method}

\subsection{Notation}
\label{sec:app-notation}
Table \ref{table:notation} presents a summary of our notation.

\begin{table}[H]
\centering
\scalebox{0.85}{
\begin{tabular}{ll}
\toprule
\textbf{Symbol} &  \textbf{Description}\\
\midrule
$t$ & the column dimension\\
$u_i$ & the unit of the $i^{th}$ row \\
$z_i$ & the label of the $i^{th}$ row \\
$v_i$ & the numeric value of the $i^{th}$ row \\
$x_i$ & the unit symbol of the $i^{th}$ row \\
$y_i$ & the characters of the $i^{th}$ row \\
$K$ & the number of possible dimensions\\
$L_t$ & the number of possible units for dimension $t$\\
$S_{u_i}$ & the number of possible unit symbols for unit $u_i$\\
\bottomrule
\end{tabular}
}
\vspace*{3mm}
\caption{A summary of the notation used by PUC.}
\label{table:notation}
\end{table}

We assume that a column of data $\textbf{y} = \{y_i\}_{i=1}^N$ consisting of $N$ rows has been read in, where each $y_i$ denotes the characters in the $i^{th}$ row. Additionally, each $y_i$ is assumed to be parsed to a numeric value $v_i$ and a unit symbol $x_i$, which may be missing for some entries, i.e., $x_i$ may be null. We propose a generative model with a set of latent variables $t$, $\textbf{u} = \{u_i\}_{i=1}^N$ and $\textbf{z} = \{z_i\}_{i=1}^N$, where $t$ denotes the dimension of a column, $u_i$ the unit and $z_i$ the label of its $i^{th}$ row. The missing and anomalous labels, denoted by $m$ and $a$ respectively, are used to model the data entries where the unit symbols are missing or anomalous. Thus, each $z_i$ can be $m$ or $a$ as well as a row unit that fit the column dimension, i.e. $z_i \in \{\text{Litre}, \text{Cubic foot}, ..., m, a\}$ given that $t$ is volume. With this noise model, we make our inference procedure robust against missing and anomalous unit symbols. 

\subsection{Parsing Unit Symbols}
\label{sec:appendix-parsing}
Unit symbols are usually positioned after quantities as in \texttt{1 L}, with some exceptions where the conventions are different. For example, they are usually placed before quantities to represent monetary amounts, e.g., \texttt{\$159000} and \texttt{\$85810}. When abbreviations are used, however, unit symbols are placed after numeric parts, e.g., \texttt{70 USD}, \texttt{19.68 AUD}. Monetary amounts can also be represented in various non-standard formats. For example, whitespace may be placed between symbols and amounts, e.g., \texttt{\$ 1012}, \texttt{\$ 964}. We develop regular expressions by taking into account possible positions of unit symbols. In particular, we can parse a numeric value expressed as a fraction, or an integer or a float, followed
by whitespace, then a string. Additionally, we remove leading and trailing whitespace as well as trailing dots. We use the following regular expression to parse a unit symbol from a given text:
\begin{Verbatim}[fontsize=\small]
import re
numeric_with_string_const_pattern = r"""
    [-+]? # optional sign
     (
         (?: \d+ \/ \d+ )   # 1/4 etc
         |
         (?: \d* [.,] \d+ ) # .1 .12 .123 etc 9.1 etc 98.1 etc
         |
         (?: \d+ \.? ) # 1. 12. 123. etc 1 12 123 etc                  
     ) ?
     # whitespace as separator
     (?: [\s]*) ?
     
     # followed by optional characters (alphanumeric, whitespace, some punctuation marks
     ([\w\s.!?\\-]*) ?     
     """
rx = re.compile(numeric_with_string_const_pattern, re.VERBOSE)
\end{Verbatim}

\subsection{Derivations for Inference}
\label{sec:appendix-derivations}

\paragraph{\bf{Column Dimension:}} Assuming that the entries of a data column are conditionally independent given the column dimension, we obtain the posterior distribution of column dimension $t$ by marginalizing over row unit and label variables $\textbf{u}$ and $\textbf{z}$ as follows:
\begin{eqnarray}\label{eqn:units-posterior-column-type}
p(t=k|\textbf{x}) &\propto& p(t=k, \textbf{x}), \nonumber \\
&=& p(t=k) \prod_{i=1}^N p(x_i | t=k), \nonumber \\
&=& p(t=k) \prod_{i=1}^N \left[ \sum_{u_i=1}^{L_k} \sum_{z_i \in \{u_i,m,a\}} p(x_i, z_i, u_i | t=k)\right], \nonumber \\
&=& p(t=k) \prod_{i=1}^N \Bigg[ \sum_{u_i=1}^{L_k} p(u_i|t=k) \Big( w_{u_i}^k p(x_i | z_i = u_i) \nonumber \\
&& \qquad \qquad \qquad \qquad  + w_{u_i}^m p(x_i | z_i = m) + w_{u_i}^a p(x_i | z_i = a) \Big)\Bigg].
\end{eqnarray}

Eq. \ref{eqn:units-posterior-column-type} can be used to estimate the column dimension $t$, since the one with maximum posterior probability is the most likely dimension corresponding to the column $\textbf{x}$. As per the equation, the model estimates the column dimension by considering all the data rows, i.e. having missing or anomalous unit symbols does not confuse the dimension inference. Note that such entries would have similar likelihoods for each column dimension, which allows the model to choose the dominant dimension for regular entries.

\paragraph{\bf{Row Label:}} Following the inference of column dimension, the posterior probabilities of each row label $z_i$ given $t=k$ and $x_i$ is obtained by marginalizing the latent unit variable $u_i$ as follows:
\begin{eqnarray}
p(z_i=j | t=k, x_i) &=& \frac{p(z_i=j, x_i | t=k)}{\sum_{\ell \in \{u_i, m, a\}} p(z_i=\ell, x_i | t=k)}, \nonumber 
\end{eqnarray}
where $p(z_i=j, x_i | t=k)$ can be calculated as follows:
\begin{eqnarray}
p(z_i=j, x_i | t=k) &=& \sum_{l=1}^{L_k} p(u_i=l, z_i=j, x_i | t=k), \nonumber \\
&=& \sum_{l=1}^{L_k} p(u_i=l|t=k) w_l^j p(x_i | z_i=j).
\end{eqnarray}

\paragraph{\bf{Row Unit:}} Given $t=k$, $z_i=j$ and $x_i$, the posterior distribution of row unit $u_i$ is obtained as:
\begin{eqnarray}
p(u_i=l | t=k, z_i=j, x_i) &=& \frac{p(u_i=l, x_i | t=k, z_i=j)}{\sum_{u_i=1}^{L_k} p(u_i=l, x_i | t=k, z_i=j)},\nonumber 
\end{eqnarray}
where $p(u_i=l, x_i | t=k, z_i=j)$ can be calculated as follows:
\begin{eqnarray}
&=& p(u_i=l|t=k) p(x_i | t=k, u_i=l, z_i=j), \nonumber \\
&=& p(u_i=l|t=k) p(x_i | z_i = j)\nonumber.
\end{eqnarray}

\paragraph{\bf{Column Unit:}} Following the column dimension inference, we set the column unit $l^*$ as follows:
\begin{equation}\label{eqn:units-posterior-column-unit}
l^* = \operatorname*{argmax}_l \sum_{i=1}^{N} p(u_i=l|t=k, x_i),
\end{equation}
where $l \in \{1,\dots,L_t\}$ denotes a possible unit for dimension $t$.

\begin{eqnarray}\label{eqn:posterior-column-unit}
p(u_i=l|t=k, x_i) &\propto& p(u_i=l|t=k)\ p(x_i|u_i=l, t=k), \nonumber \\
&=& p(u_i=l|t=k)  \Bigg[ \sum_{\ell \in \{u_i, m, a\}} \Big( p(x_i,z_i=\ell|u_i=l, t=k)\Big) \Bigg], \nonumber\\
&=& p(u_i=l|t=k)  \Bigg[ \sum_{\ell \in \{u_i, m, a\}} \Big( p(z_i=\ell|u_i=l)\ p(x_i|z_i=\ell) \Big) \Bigg]\nonumber.
\end{eqnarray}

\paragraph{\bf{Correcting Anomalous Unit Symbols:}} Row label inference annotates each data entry either as a unit, missing or anomalous. We assume that the units of anomalous entries are encoded by anomalous unit symbols (e.g., \texttt{ltrs} for \texttt{litres}) and can be identified by mapping anomalous unit symbols (e.g., \texttt{ltrs}) to known unit symbols (e.g., \texttt{lt}) based on the edit-distance \cite{levenshtein1966binary}. The edit-distance measures the minimum number of operations (addition, deletion or substitution) that needs to be done to transform a string to another, and can handle misspellings and non-standard abbreviations. Note that we restrict the set of unit symbols to be compared with according to the column dimension, i.e., we compare \texttt{ltrs} with known unit symbols that encode units of volume when the column dimension is inferred as volume.

\paragraph{\bf{Canonicalizing Units:}} Following the inference of the row units and the column unit, we are now interested in representing each row with the same unit by scaling its numerical value (e.g., converting the data entry \texttt{1 m} to \texttt{100 cm} when the column and row units are respectively centimetres and metres). Currently, we convert units via an existing tool named Pint \cite{pint2019}. See Sec.\ \ref{sec:related_work} for a detailed discussion of Pint.

\section{Additional Experimental Results}
\label{sec:app-exp-res}
\subsection{Dimension Inference}
PUC correctly predicts the dimensions of all data columns, except one data column for which all the competitor methods also fail. These failures result from the inability to parse the data values such as \texttt{5\textquotesingle10\textquotedbl} where \texttt{\textquotesingle} and \texttt{\textquotedbl} denote respectively feet and inches. We could improve our regular expression to parse such formats, which we have not done in order not to optimise on test datasets. On the other hand, the leading competitor methods are S-NER and Quantulum, which achieve the same overall accuracy. They both fail to identify the column dimensions of seven data columns. To compare our method with these baselines, we present their normalised confusion matrices in Fig.\ \ref{fig:units-confusion-matrices-explicit}, normalised so that a column sums to 1. 

\begin{figure}[ht!]
	\centering	
	\begin{subfigure}[b]{0.32\textwidth}
    \centering
		\includegraphics[width=\textwidth]{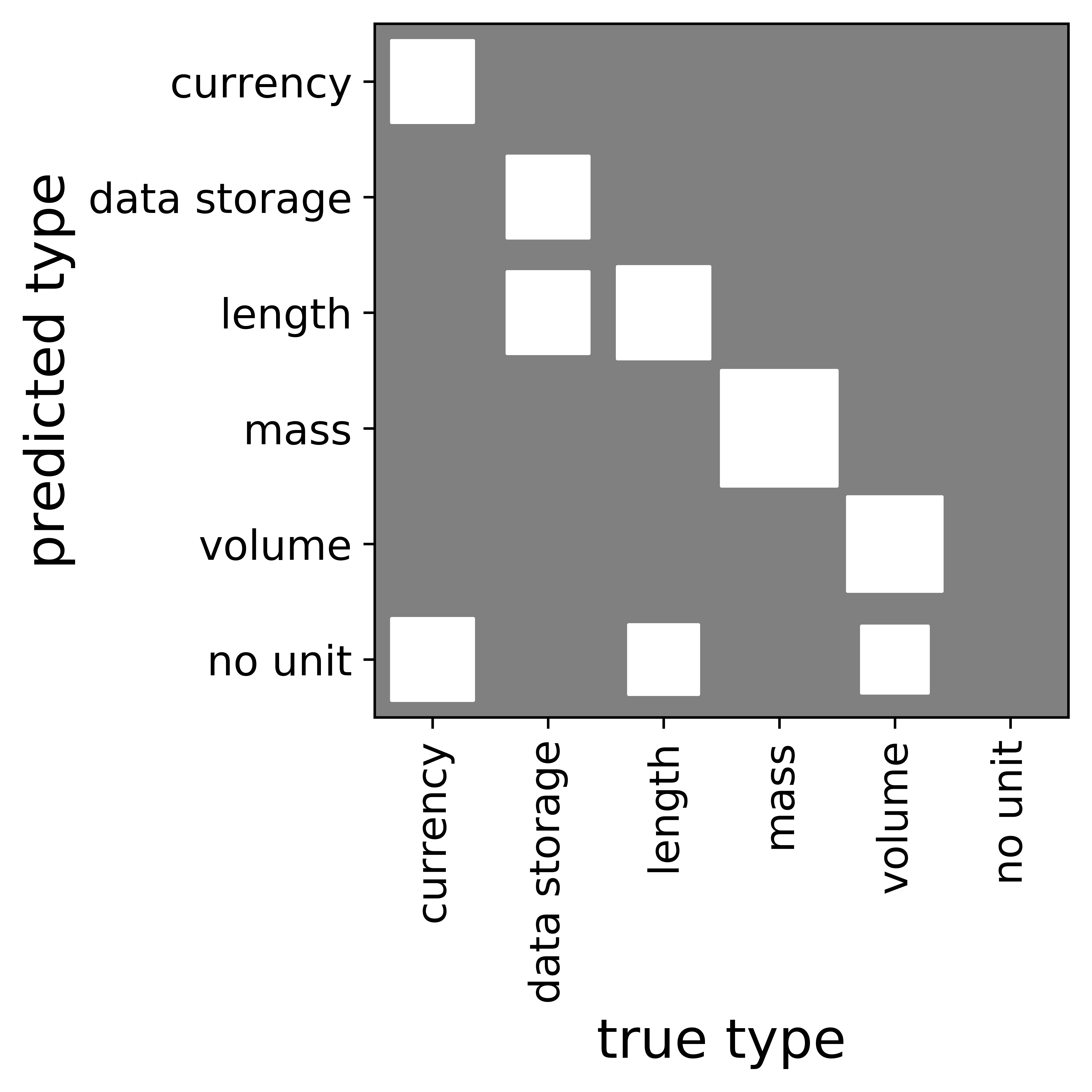}
		\caption{Quantulum}
	\end{subfigure}%
	\begin{subfigure}[b]{0.32\textwidth}
    \centering
		\includegraphics[width=\textwidth]{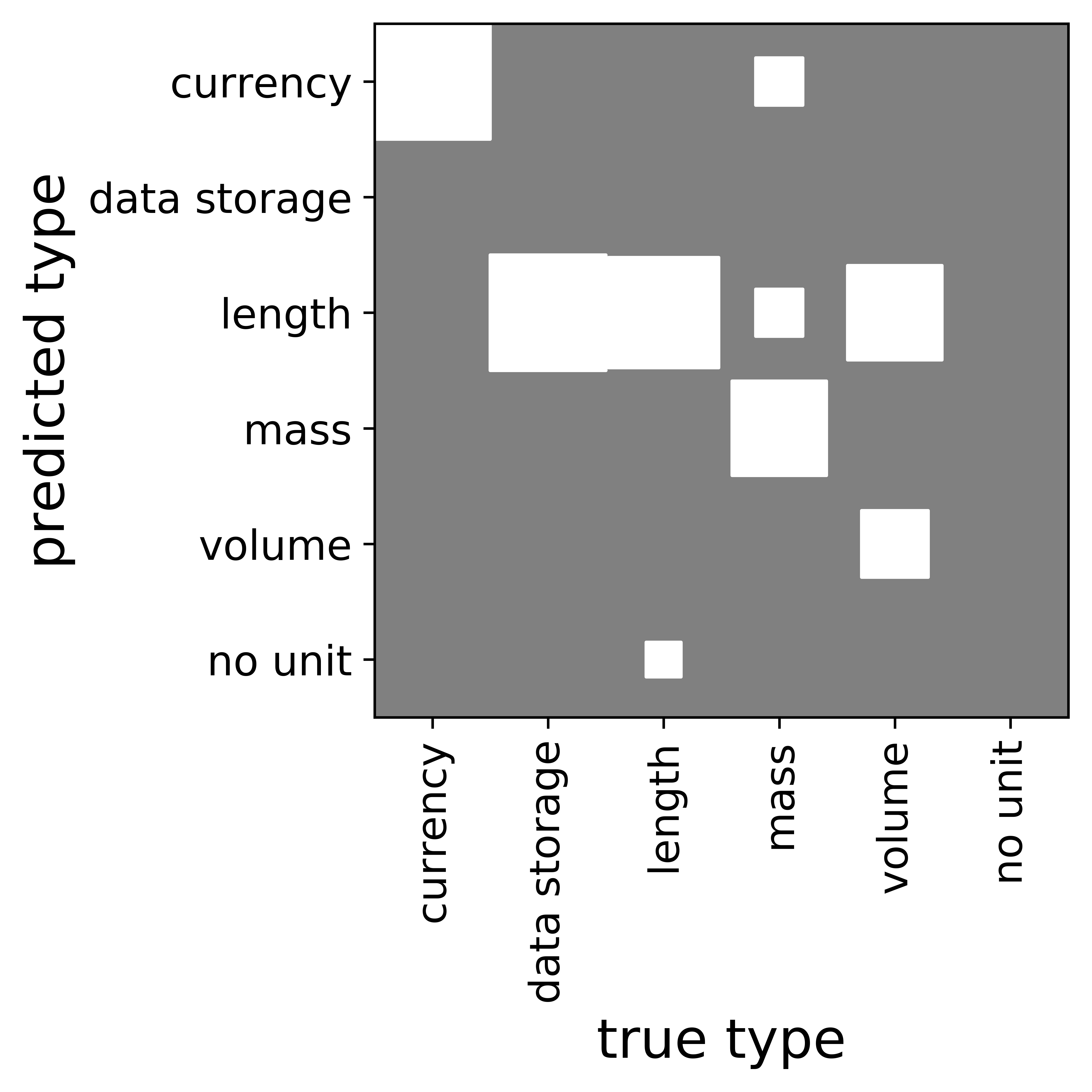}
		\caption{S-NER}
	\end{subfigure}%
	\begin{subfigure}[b]{0.32\textwidth}
    \centering
		\includegraphics[width=\textwidth]{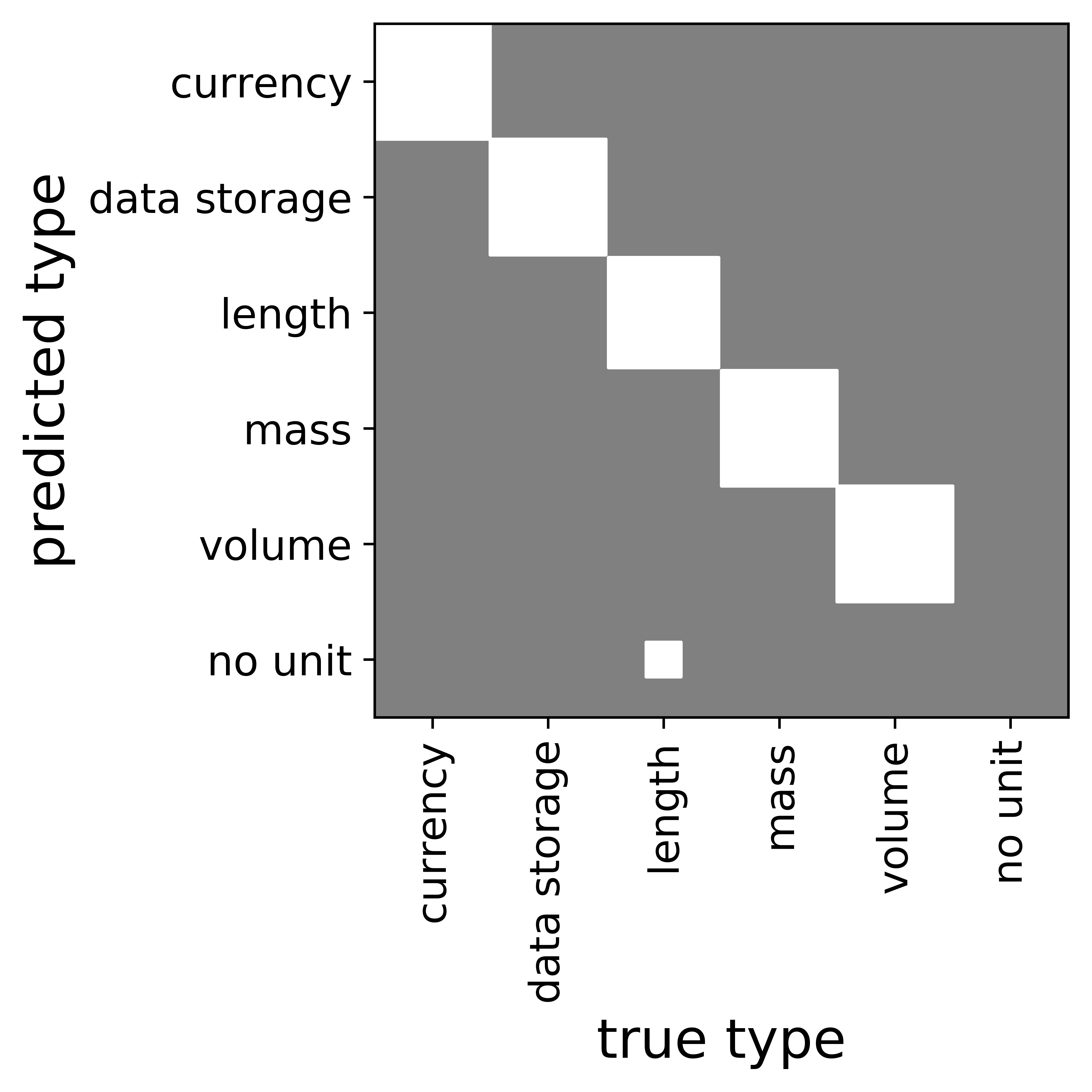}
		\caption{PUC}
	\end{subfigure}
	\caption{Normalized confusion matrices for (a) Quantulum, (b) S-NER and (c) PUC plotted
as Hinton diagrams, where the area of a square is proportional to the
magnitude of the entry.}
	\label{fig:units-confusion-matrices-explicit}
\end{figure}

Fig.\ \ref{fig:units-confusion-matrices-explicit}(b) shows that S-NER tends to infer the column dimension as length. These failures can be explained by the differences in the number of characters of unit symbols. S-NER performs better on longer unit symbols, which may not be surprising as it uses n-grams for feature extraction. For example, the confusions between volume and length occur on the HES dataset, where the dimensions of units symbols for cubic foot (e.g., \texttt{cuft}, \texttt{cu.ft}) are correctly predicted as volume. Nevertheless, the dimensions of unit symbols for litre (e.g., \texttt{l}, \texttt{L}) are predicted as length instead of volume, which result in incorrect predictions since they are more frequent in the data. Character-level features (e.g., n-grams) could be useful to handle variations in the data such as misspellings; however, they may lead to limited performance on short unit symbols. This result indicates the advantage of incorporating knowledge about unit symbols directly into the model, as in PUC. 

As we discuss above, Quantulum cannot parse measurements such as \texttt{5\textquotesingle10\textquotedbl} and fails to identify the dimension of the corresponding data column. The confusion between length and data storage occurs in a data column where kilobyte and megabyte are respectively encoded by \texttt{K} and \texttt{M}. Here, Quantulum mislabels \texttt{M} as metre and does not generate a prediction for \texttt{K}. A detailed inspection of the remaining five columns shows that numeric values are missing in the entries, which cannot be handled by Quantulum.  

To determine whether the column dimension predictions of PUC and the leading competitor methods (namely S-NER and Quantulum) are significantly different, we apply a variation of the McNemar's test as the number of samples is low (see e.g., \cite{edwards1948note}). This test assumes that the two methods should have the same error rate under the null hypothesis. We compute the exact p-value $2 \sum_{i=n_{10}}^{n_{01}} {n\choose i} 0.5^i (1-0.5)^{n-i}$ where $n=n_{01}+n_{10}$ with $n_{01}$ and $n_{10}$ which respectively denote the number of columns misclassified by only a competitor method, and by only PUC. The test to compare PUC and S-NER results in a p-value of 0.03 since $n_{01}$ and $n_{10}$ are respectively equal to 6 and 0. These results reject the null hypothesis that the means are equal and confirm that they are significantly different at the 0.05 level. We obtain the same result from the test between PUC and Quantulum, as $n_{01}$ and $n_{10}$ are the same as S-NER, which confirms that they are significantly different at the 0.05 level.

Lastly, we have evaluated the runtime of each method per data column. Figure \ref{fig:units-runtimes} shows that S-NER is the slowest method, whereas PUC is the fastest on average. The leading competitor method is Pint; however, the variation in its runtime is higher than PUC. Note that we have not explicitly optimised our method for speed, which may improve further its scalability.

\begin{figure}[ht!]
\centering	
\includegraphics[width=.7\textwidth]{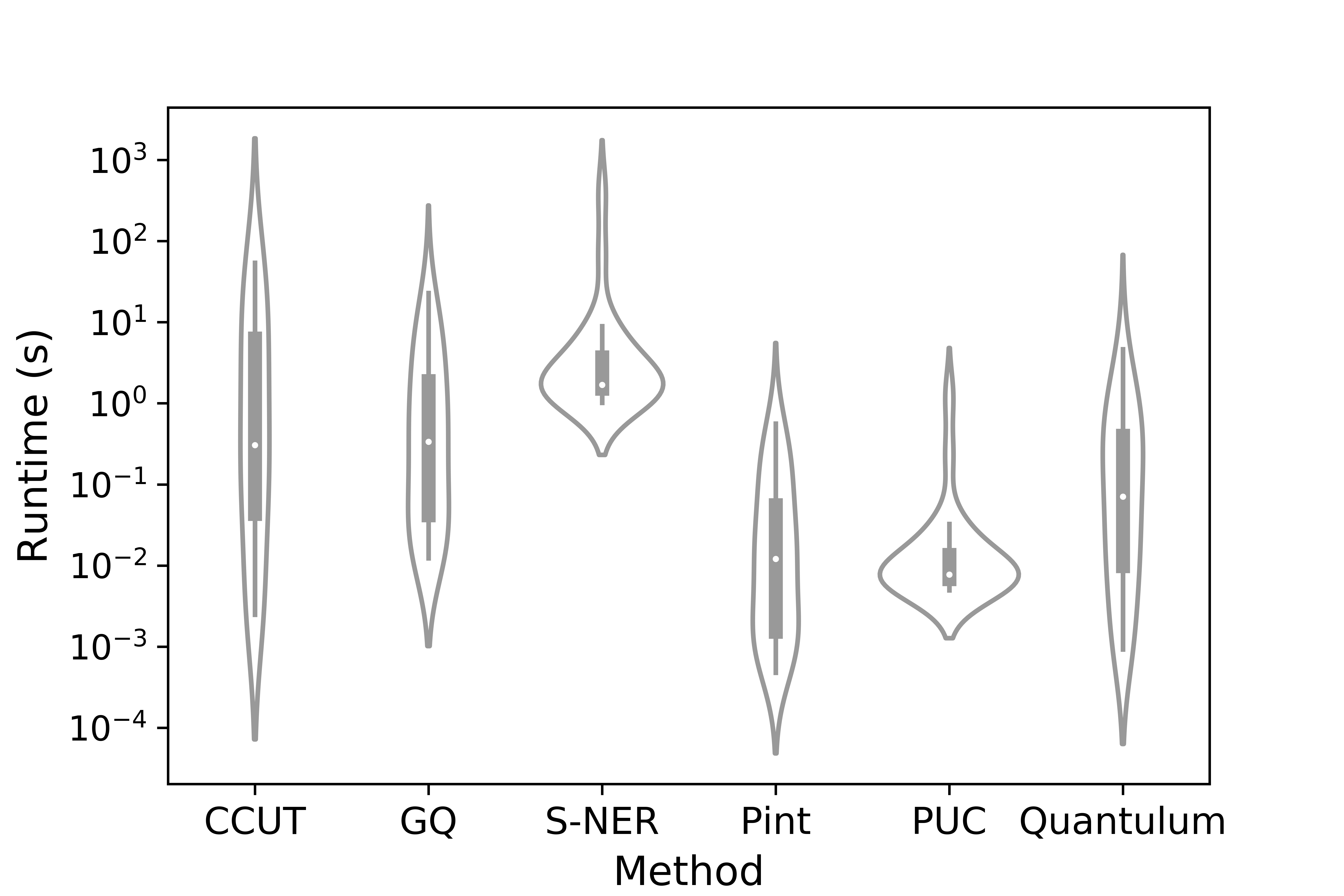}		
\caption{Runtime violin plots denote the time in seconds taken to infer dimensions per column. The dot, box, and whiskers respectively denote the median, interquartile range, and 95\% confidence interval.}
\label{fig:units-runtimes}
\end{figure}

\subsection{Unit Identification}
PUC outperforms the competitor methods by a large margin on 4 datasets  (143\dots23, HES, MBA and Zomato) out of the 15. Note that we exclude the Taser dataset from the evaluations since none of the methods could parse its values such as \texttt{5\textquotesingle10\textquotedbl} where \texttt{\textquotesingle} and \texttt{\textquotedbl} denote respectively feet and inches. On the remaining 11 datasets, there is at least one competitor method competitive with ours. 

The performance gap between PUC and the competitor methods reflects the importance of mapping anomalous unit symbols to known symbols through string-similarity. For example, on the HES dataset, the competitor methods could accurately identify only a few unit symbols, whereas our method could successfully identify almost all of the unit symbols. Out of 14 unique unit symbols, CCUT identified \texttt{L} and \texttt{cuft}, and Pint identified \texttt{L}, \texttt{l} and \texttt{litres}. In addition to these three unit symbols, Quantulum identified \texttt{Litres}. Surprisingly, GQ, which is one of the state-of-the-art methods in identifying units in text documents, could not identify any of these unit symbols. PUC, on the other hand, could identify 12 unit symbols correctly, with only two unidentified unit symbols (\texttt{Cu} and \texttt{cf}). 

We observe that Quantulum performs better than PUC on the Arabica dataset. This result is mainly due to the Altitude column where metre is encoded by \texttt{M}, which is a known symbol for mile. Consequently, our method predicts the units of such entries as mile rather than metre. We could avoid this confusion by making row units dependent, so that the presence of unit symbols (e.g., m, metres) in the other data entries that encode the same unit (e.g., metre) is treated as an indicator of \texttt{M} being a symbol for metre rather than mile. Here, we do not adapt our model accordingly so that it is not optimised on test datasets. 

On the Zomato dataset, PUC performs better than the competitor methods. However, it still cannot identify 4 symbols out of 10 currency symbols and maps them to incorrect units based on string similarities. For example, PUC identifies \texttt{QR} (a symbol for Qatari Riyal) as an anomalous unit symbol. As \texttt{QR} is an out-of-dictionary unit symbol, it is mapped by PUC to the known BRL, which is a symbol for the South African Rand. Although this mapping mechanism can improve performance in certain cases (e.g., the HES dataset), it can sometimes fail.

To determine whether the performances of PUC and Pint are significantly different, we have applied a paired t-test on the differences of the accuracies, i.e., Accuracy(PUC) - Accuracy(competitor method). We have calculated the t-statistic of 4.85 and the p-value of 0.0002 for CCUT, the t-statistic of 8.61 and the p-value of 0.000001 for GQ, the t-statistic of 2.74 and the p-value of 0.01 for Pint and the t-statistic of 2.55 and the p-value of 0.02 for Quantulum. These results reject the null hypothesis that the means are equal and confirm that they are significantly different at the 0.05 level.

Lastly, Table \ref{table:accuracy-unit-identification} indicates that Pint and Quantulum perform best among the competitor methods. We have compared their performances through a paired t-test, i.e., Accuracy(Quantulum) - Accuracy(Pint), which suggests that they are not significantly different at the 0.05 level (the t-statistics of 0.18 and the p-value of 0.86). This result suggests that neither method is comprehensive enough to handle various unit symbols observed in different datasets. 

\end{document}